# Survival of the fastest — algorithm-guided evolution of light-powered underwater microrobots

Mikołaj Rogóż[1*], Zofia Dziekan[1], Piotr Wasylczyk[1]

[1]Photonic Nanostructure Facility, Faculty of Physics, University of Warsaw, Pasteura 5, 02-093, Warsaw, Poland

*e-mail: mikolajrogoz@uw.edu.pl

**Depending on multiple parameters, soft robots can exhibit different modes of locomotion that are difficult to model numerically. As a result, improving their performance is complex, especially in small-scale systems characterized by low Reynolds numbers, when multiple aero- and hydrodynamical processes influence their movement. In this work, we optimize light-powered millimetre-scale underwater swimmer locomotion by applying experimental results – measured swimming speed – as the fitness function in two evolutionary algorithms: particle swarm optimization and genetic algorithm. As these soft, light-powered robots with different characteristics (phenotypes) can be fabricated quickly, they provide a great platform for optimisation experiments, using many competing robots to improve swimming speed over consecutive generations. Interestingly, just like in natural evolution, unexpected gene combinations led to surprisingly good results, including eight-fold increase in speed or the discovery of a self-oscillating underwater locomotion mode.**

Liquid crystal elastomers (LCEs) are smart materials, capable of reversible actuation in response to external stimuli such as electric field, heat or light[1]. This behaviour arises from their unique structure, which combines the anisotropy of liquid crystals with the elasticity of loosely cross-linked polymer chains, while the type of deformation can be pre-programmed into the material during the fabrication process, through various alignment methods[2]. These characteristics make LCEs promising for applications in soft robotics[3–5], where flexibility, adaptability, and responsiveness are crucial. In robotics, LCEs have been used to create soft actuators that mimic the natural movements of biological organisms[6–8], offering a pathway to lifelike, efficient robotic systems. Here, we present swimming robots inspired by our previous work on caterpillar-like crawling robots[9]. The robot motion is driven by periodic, travelling deformations in the LCE body, induced by a scanned laser beam (Fig. 1). Several key parameters influence the robot speed, including laser power, scanning frequency, and the geometry of the robots' body. As LCE deformation can also be influenced by light polarization[10], it was included as one of the parameters.

Optimising the performance of such robots is a challenging task, often addressed through simulations that predict the robot's behaviour based on its design parameters[11]. Despite significant efforts to align simulations with experimental outcomes, they can generate solutions that, when implemented, fail to perform as expected in real-world conditions[4,12–15]. This discrepancy arises from the challenge of accurately modelling the complex interactions between soft materials and their environment. This issue is especially pronounced in small-scale systems, where factors like local variations in a medium viscosity, fluid flow patterns or heat dissipation make precise modelling difficult[16,17].

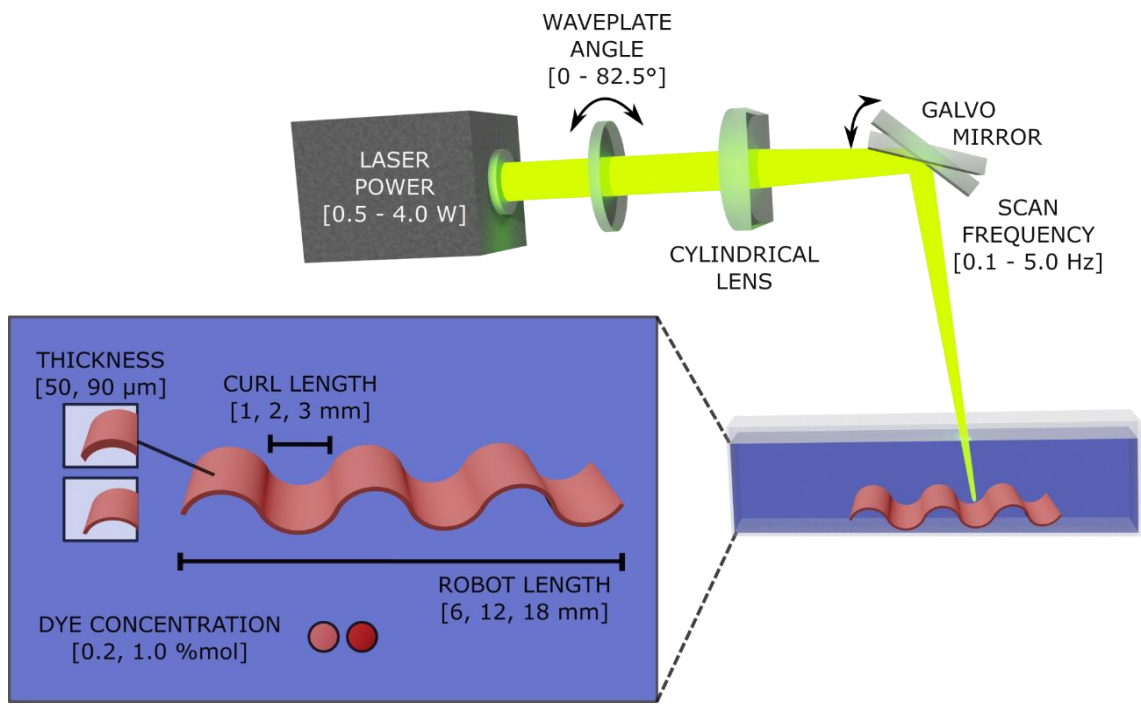

**Fig. 1 Parameters (genes) characterizing the robot and the setup used for "robot races" – swimming speed measurement.** LCE robots were submerged in a narrow, water-filled tank and actuated by heat generated through laser light absorption. To achieve underwater locomotion, the laser scan in one direction was fast enough to avoid inducing a response in the material, while the scan in the opposite direction was slow enough to generate a deformation traveling along the robot. The setup allowed for control over (1) laser power, (2) scanning frequency, and (3) polarization direction. Other optimized parameters included the robot body characteristics: (4) film thickness, (5) length, (6) curl length, (7) tail direction (up or down), and (8) dye concentration, all of which influenced the underwater swimming dynamics.

Therefore, our optimisation approach diverges from the reliance on simulations, where the performance of individual machines is modelled. Instead, we fabricate several sets (generations) of robots and test them in their intended working environment. Rather than relying on simulated performance metrics, we used iterative algorithms where robots raced against each other to identify the fastest swimmers out of 345,600 possible parameter combinations. Given the vast search space, we selected particle swarm optimization (PSO) and genetic algorithm (GA) due to their efficiency in exploring high-dimensional, complex systems without requiring an explicit mathematical model. These algorithms effectively integrate experimental feedback, using measured robot speeds as fitness values to refine designs iteratively. GA promotes diversity through crossover and mutation, preventing premature convergence[18], while PSO balances local and global search strategies to optimize performance efficiently[19]. These algorithms then generated the parameters for the next generation of robots, which competed in subsequent races (Fig. 2). This approach of integrating experimental measurements directly into optimization algorithms – often referred to as data-driven computing[20] or feedback optimisation[21] – has been so far applied across various scientific disciplines. However, its application to complex systems like swimming robots remains relatively rare, particularly in achieving significant performance gains within a limited number of iterations. This suggests that further exploration of such optimization techniques could enhance the efficiency and adaptability of these systems.

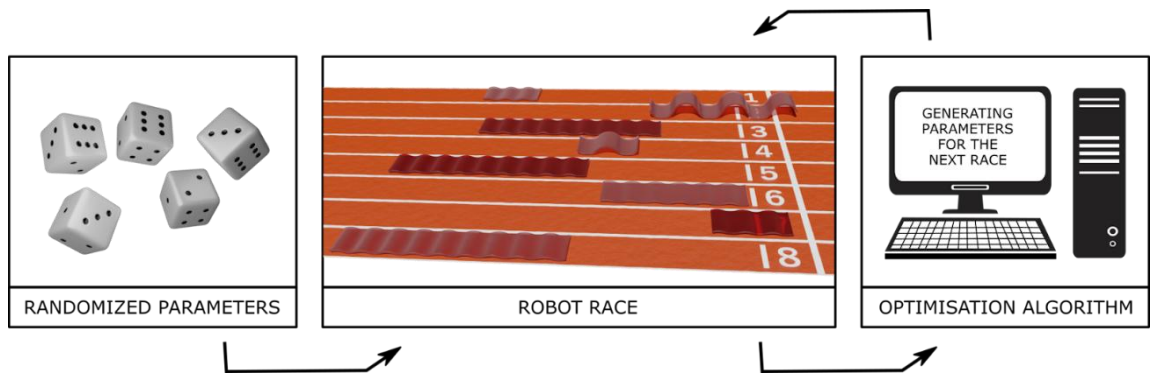

**Fig. 2 Optimisation of the soft microrobots in the search for the fastest swimmers.** The iterative optimisation involved eight sets of parameters (genotypes) – and the corresponding eight robots (phenotypes) – per generation Robot parameters in the initial generation were randomized, while in subsequent generations, they were chosen by one of the studied optimization algorithms based on the experimentally measured fitness function – swimming speed – measured experimentally.

## RESULTS

**Robot Parameter Space**

The soft robots were made from LCE films capable of light-induced, temperature-mediated deformation from flat into an accordion-like shape[22]. Each film consisted of alternating twisted nematic domains, each rotated by 90° relative to its neighbours. The width of these domains defined the parameter referred to as "curl length" (Fig. 1), which corresponds to the width of the slits in the mask applied during the fabrication process[23].

Based on preliminary experiments, we identified eight key parameters that most significantly influence motion, defining both the robot's shape and its powering system (Fig. 1). The material properties were determined by a combination of thickness (50 or 90 µm), dye content (0.2 or 1.0 mol%), and curl length (1, 2, or 3 mm). The robot lengths were chosen to be multiples of the curl length (6, 12, or 18 mm) to create areas with consistent bending direction within each film. Details of robot fabrication can be found in the Materials and Methods section. Another important parameter was the orientation of the film relative to the laser beam. To determine this, we marked one of the shorter edges of the film with a permanent marker and defined the "tail direction" as 1 when the marked edge bent upwards at room temperature, and 0 otherwise. The marked edge of the film was consistently positioned on the same side of the system.

In the optical setup, the beam scanning frequency was set between 0.1 to 5 Hz with 0.1 Hz increments. Preliminary experiments revealed that frequencies above 5 Hz induced stationary deformation of the LCE film[24], due to quasi-continuous illumination, resulting in no body movement. The laser power was chosen within the range of 0.4 to 3.2 W, with 0.4 W increments. Additionally, a half-wave plate was used to rotate the linearly polarized laser beam, allowing the angle between the polarization direction and the scan direction to be adjusted from 0° to 165° in 15° increments.

**Optimisation Algorithms Parameters**

To optimise robot performance, we implemented a five-iteration process, starting with a randomized initial generation followed by four optimisation cycles, each evaluating the performance of eight robots. This optimisation spanned an 8-dimensional parameter space, with each dimension representing a key robot/system characteristic. However, given that GA and PSO have tunable

parameters – such as mutation chance in GA and acceleration coefficients in PSO – before venturing into experiments, we ran computer simulations to determine the most effective algorithmic variations and parameter settings. The details of these simulations and their results are described in the Materials and Methods section.

The first approach, using a GA[25], modifies a group of candidate solutions, called "individuals" or "phenotypes" in each iteration, starting with a randomized initial population. Each individual consists of variables known as genes, which, in our case, represent the parameters of the robots and scanning system. A single robot, powered by specific laser parameters, defines the phenotype. New solutions (children) are typically generated by combining parents from the previous generation through crossover. Additionally, each child's genetic material may mutate, randomly altering one or more genes. Various GA variations exist, differing in parent selection methods, gene combination techniques, and mutation rates.

In our GA implementation, we first selected four pairs of parents, allowing for duplicates in the selection process. However, within each pair, two parents were always different, meaning no individual is paired with its twin. A single mutation was then applied to a randomly chosen gene of each parent. New individuals were created through uniform crossover, each child inherited half of its genes from one parent and the rest from the other. The unused genes are passed to the second child. Identical individuals were not allowed; any duplicates underwent a single random gene mutation to ensure a unique genotype.

We employed the rank selection method with an elitism-inspired approach, where selection was performed among the top eight individuals found throughout the entire optimization process rather than only considering individuals from the previous generation. In rank selection, the probability of selection was determined by the fitness rank within the population:

$$P(k) = B(1-B)^k + (1-B)^n/n \tag{1}$$

where $n$ is the number of individuals participating in the selection, $B = n^{-2/3}$ and $k$ represents the individual's position in the ranking (starting with 0).

The second optimisation method used a PSO[26] – an algorithm inspired by the collective behaviours observed in nature, such as birds flocking or fish schooling. In PSO, a group of potential solutions, referred to as "particles" (analogous to phenotypes in GA), explores the searched space. Each particle represents a point in this space and iteratively adjusts its position to move closer to an optimal solution. The position of each particle is updated by calculating its velocity, which determines the direction and magnitude of its movement. This velocity is influenced by a combination of factors: the particle's current velocity, the distance to its personal best position (the best solution it has found so far), and the distance to the global best position (the best solution found by any particle in the swarm), determined using the formulas[26]:

$$v^{i+i} = wv^i + c_1 r_1 (x_{pbest} - x^i) + c_2 r_2 (x_{gbest} - x^i), \tag{2}$$

$$x^{i+1} = x^i + v^{i+1}, \tag{3}$$

where: $x^i$ is the position of the particle in the i-th iteration, $v^i$ is the current velocity, $w$ is the inertia weight that influences how much of the previous velocity is retained, $c_1$ is the cognitive coefficient, guiding the particle towards its personal best position, $c_2$ is the social coefficient, guiding the particle towards the global best position, $x_{pbest}$ is the personal best position of the particle, $x_{gbest}$ is the global best position found by the swarm, $r_1$ and $r_2$ are random numbers between 0 and 1, introducing stochasticity to the process. We used the following values: 0 for the inertia weight, 0.2 for the cognitive coefficient, and 1.4 for the social coefficient. Each particle coordinates were rounded to the nearest value in the robot parameter space, except for the polarization angle, where periodic boundary conditions were applied. Additionally, if any component of a particle's velocity exceeded the allowed range for a given parameter, it was limited to the range defined by the difference between its maximum and minimum values.

Duplicates of candidate solutions within the swarm were prohibited across all iterations, except when they occurred within the same iteration. If duplicates appeared, additional PSO steps were performed. Sometimes PSO steps could not eliminate duplicates from the swarm. If a duplicate remained after five PSO steps, a single random gene mutation, similar to the GA, was applied until a new individual was generated. When a particle reached a new global maximum, its velocity was calculated by multiplying a random number between -1 and 1 by a vector whose components were equal to the differences between the maximum and minimum values of each parameter.

**Results of the Data-Driven Optimisation**

We conducted four optimization experiments using GA and PSO twice, each starting from two different sets of randomized parameters. The robot fabrication and measurement procedures are detailed in the Materials and Methods section, while the final results are presented in Fig. 3 and Supplementary Fig. 1. Both algorithms helped identify parameter combinations that failed to induce movement, providing valuable information about the system. Precisely, in the first generation, 50% of robots remained nearly stationary, with speeds below 0.1 cm/min (Fig. 3).

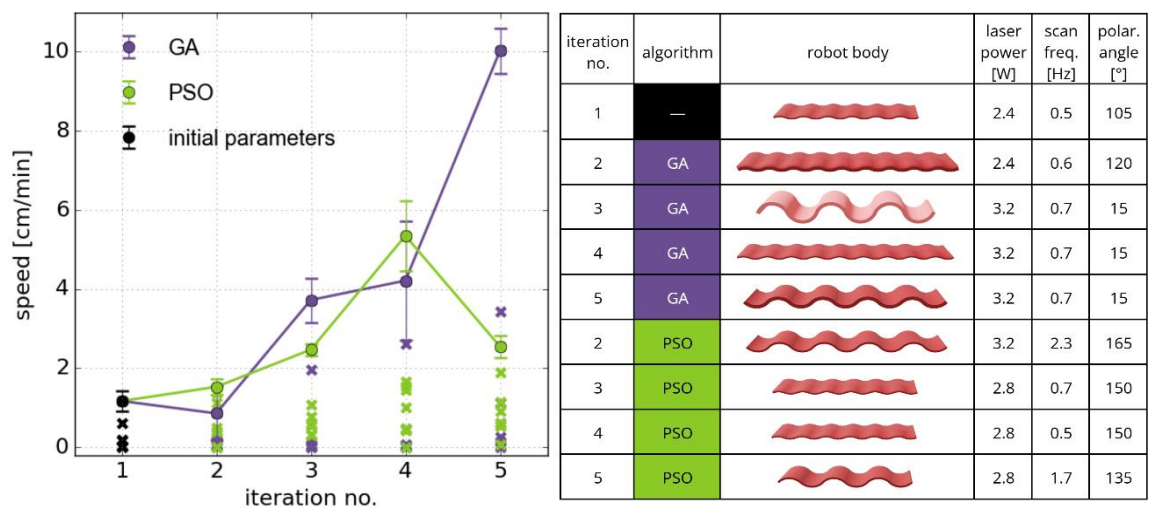

**Fig. 3 Speeds of soft swimming microrobots achieved in consecutive iterations of the first optimization experiment. (Left)** Measured speed of the robots, fabricated based on the parameters generated by the GA and PSO. Dots represent the best individuals in each iteration, while bars indicate the standard deviation from their speed measurements. X's correspond to the remaining robots from the same generation. **(Right)** Table summarising the parameters of the fastest swimmers from each iteration, for both optimization methods. The robot body shape is illustrated with a 3D model, and the laser parameters are listed alongside.

The optimization process led to significant performance improvements, increasing the robot maximum speed three to eight times over five generations, highlighting the effectiveness of our approach (Supplementary Video 1). The initial, randomly generated robots exhibited a broad range of swimming speeds, with the fastest reaching 1.2 cm/min in the first trial and 1.7 cm/min in the second. After four optimization iterations, the best-performing robots achieved speeds of 10.0 cm/min for GA and 5.3 cm/min for PSO in the first experiment (Fig. 3). In another, GA and PSO reached 5.0 cm/min and 4.9 cm/min, respectively (Supplementary Fig. 1). These results confirm that optimization using real-world measurements as a fitness function can significantly enhance robot performance in just a few iterations. Interestingly, despite starting with a lower initial maximum speed, the first experiment ultimately achieved higher final speeds than the second. This suggests that the starting point had minimal influence on optimization effectiveness and that the evolutionary approach effectively guided the system toward improved solutions.

Although both GA and PSO effectively increased the robot speed, their optimization behaviors differed. GA consistently produced higher-performing robots, especially in the first optimization trial, where it outperformed PSO by nearly a factor of two. In contrast, PSO often pushed parameters toward boundary values, sometimes resulting in less efficient configurations. It frequently converged to extreme scan frequencies and laser powers, which, while occasionally beneficial, did not always yield the best results. PSO's faster convergence made it efficient in early iterations but less reliable in consistently identifying the best-performing designs. However, the minimal difference between the final GA and PSO results in the second trial suggests that the parameter range for the fastest robots – those exceeding 10 cm/min – is very narrow. This implies that the variation in algorithm performance may be more due to randomness than a clear advantage of one method over the other.

The most unexpected outcome of the optimization was the emergence of a self-oscillating mode of locomotion, discovered during PSO runs. While self-oscillations of LCE films have been studied by several researchers[27–29], there are limited reports focusing on this specific mechanism of swimming locomotion[30]. This mode was observed when the algorithm selected low scan frequencies (0.1 Hz), high laser power, and 3 mm curl length – a combination that led to continuous oscillations in a localized region of the robot's body. These oscillations synchronized with the laser beam movement, allowing the robot to closely follow the beam with minimal delay (Supplementary Video 2). While self-oscillations did not produce the highest speeds, they enabled robots to move in a highly controlled and repeatable manner. This suggests that such locomotion mode could be leveraged for precise navigation, potentially improving the maneuverability of soft robots in applications where directional control is as important as speed.

The self-oscillations never persisted for the full laser scan. The robot consistently exited the oscillation mode before the scan ended and began to jump with random displacements during subsequent scans. We attempted to manually identify parameters that would allow the robot to be propelled by oscillations along the length of the experimental container, achieving slightly larger distances (Supplementary Video 3). Most likely, the oscillations break because specific conditions necessary to maintain them are related to the angle between the laser beam and the LCE film and thus are not consistently satisfied throughout the full scan performed by the experimental setup. The angle changes slightly across the container due to the underwater positioning of the robot, which disrupts sustained oscillations.

## DISCUSSION

The data-driven evolutionary approach offers a practical alternative to purely simulation-based methods, particularly for complex physical systems where precise modelling is challenging. Our study demonstrates that combining experimental feedback with optimization algorithms is a powerful strategy for improving the performance of soft robots. By directly integrating experimental results as fitness functions in genetic algorithm and particle swarm optimization, we effectively improved the swimming speed of light-driven soft robots. Across 144 evaluated robots, the algorithms have significantly improved their performance in just four iterations and revealed various modes of locomotion. The fastest recorded robot achieved a speed of 10 cm/min, corresponding to 0.09 body lengths per second (BLs⁻¹). Compared to other underwater swimming robots based on smart material actuation[31], this is a surprisingly good result, especially considering the simplicity of the robot body.

Our findings highlight key differences between the studied algorithms. While both GA and PSO successfully enhanced performance, PSO's tendency to reach boundary conditions made it more likely to explore unconventional solutions, such as self-oscillations. In contrast, GA exhibited more stable and consistent performance improvements across iterations. These insights suggest that refining PSO's parameter space, potentially by adjusting its social and cognitive coefficients, could mitigate its bias toward extreme values.

PSO's high social coefficient (1.4, see Materials and Methods) promotes convergence toward global optima, likely causing its tendency to favor low scan frequencies and high laser powers. As a result, a self-oscillating mode of locomotion emerged during the PSO runs, where a small region of the robot body exhibited continuous oscillation, moving synchronously with the scanning beam. While not the fastest locomotion mode, this self-oscillating behavior presents a unique opportunity for precise robot control, as then it moves, closely following the laser beam.

Our study not only demonstrates the effectiveness of evolutionary optimization in soft robotics, but also highlights the value of real-world experimental feedback in discovering novel locomotion strategies. The emergence of self-oscillating motion, an outcome that would have been difficult to predict through conventional simulations, underscores the potential of this approach to uncover unexpected yet functional behaviors. The ability to iteratively refine robot designs through direct experimentation, without relying on complex theoretical models, makes this method highly scalable and adaptable to a wide range of robotic applications. Future research can build upon this foundation by extending the method to other bio-inspired robotic systems, leveraging additional optimization techniques, and exploring new applications where adaptability, precision, and efficiency are crucial.

## MATERIALS AND METHODS

### Selection of Optimization Algorithm Parameters

Simulations were performed to determine the optimal algorithm parameters. The fitness function was defined as the sum of one-dimensional normal distributions:

$$f(x) = \sum_i^8 \frac{exp\left(-\frac{1}{2}A\right)^2}{\sigma\sqrt{2\pi}}, A = \frac{x_i - 0.75(x_{imax} - x_{imin}) - x_{imin}}{\sigma}, \quad (4)$$

where: $x$ is an 8-dimensional vector that represents the parameters of a given robot, $x_i$ is the i-th coordinate of the vector $x$, $x_{imax}/x_{imin}$ is the maximum/minimum value of the $x_i$, while σ is the standard deviation of the distribution. The available values for each coordinate correspond to the robot parameter space. The peak of the normal distributions was shifted to 0.75 of the value range for the given parameter to introduce differences in fitness function values for dimensions where parameters take binary values. Although this formula does not accurately describe the speed of our robots, it can provide qualitative insights into the optimization of a multidimensional, quantized function. A similar fitness function was reported in [32], but in that case, the fitness was determined by the maximum rather than the sum.

We conducted simulations using four different values of σ: 0.05, 0.1, 0.25, and 0.5. The first iteration was randomized, followed by four additional iterations aimed at maximizing the fitness function value.

This process was repeated 1,000 times for every variation of the algorithm. To analyse the results, we compared the average of the highest fitness function values achieved in each simulation.

To assess optimization performance, we systematically tested all combinations of selection methods and mutation rates. Specifically, we implemented an elitism-inspired strategy and compared it with selection among all individuals from previous iterations, allowing up to 24 individuals to be chosen by the fourth iteration. Additionally, we evaluated two parent selection strategies: roulette wheel, where selection probability is proportional to the robot's fitness, and rank, where probability is determined by the fitness rank within the population, as described by formula (1).

To maintain genetic diversity in the small population, we applied a high mutation rate, allowing more than one gene per individual to be mutated. The mutation rate could exceed 1, which we interpreted as follows: if m is the mutation rate value, we applied mutations to $\lfloor m \rfloor$ genes, with an additional mutation occurring with probability $m - \lfloor m \rfloor$. For instance, a rate of 2.4 meant that 2 genes were always mutated, with a 40% probability of mutating a third gene. Mutations were applied to randomly selected genes and affected both parents before crossover.

We also investigated an adaptive mutation rate[18], where the mutation probability depends on the fitness value, described by the formula:

$$m(f) = (f_{max} - f)\frac{m_{max} - m_{min}}{f_{max}} + m_{min} \qquad (5)$$

where $f_{max}$ is the highest fitness value recorded during the optimization, $m_{max}$ is the maximum mutation rate, and $m_{min}$ is the minimum mutation rate. Simulations were performed for all combinations of $m_{min}$ (0 to 2) and $m_{max}$ (0 to 3) in 0.1 increments, ensuring that $m_{max} \geq m_{min}$.

In the case of PSO simulations, we examined all combinations of the inertia weight, cognitive coefficient, and social coefficient in the range from 0 to 3 with a step of 0.2. When a particle reached a new global maximum with an inertia weight of 0, its velocity was determined by scaling a random number (-1 to 1) with a vector representing the parameter range.

Simulations of the GA performance indicated that the elitism-inspired selection outperformed selection among all individuals measured in previous iterations regardless of the σ value, in the fitness function. The roulette selection method was superior only for σ = 0.05; for other values, rank selection produced better results (Fig. 4). σ values of 0.1 or 0.25 corresponded better to our previous experience with light-driven robots[9,33], where the range of laser power and scan speed associated with higher robot speeds was relatively wide. Therefore, we selected specific algorithm variations for our evolutionary experiments. For the GA, we employed the elitism-inspired approach with the rank selection method. Low σ correlated with better performance of higher mutation rate. Specifically, for σ = 0.05, the best results with an adaptive mutation rate were obtained, where the minimum rate was usually above one and the maximum above two. For other σ values, both minimum and maximum rates performed best close to one with no significant advantage of adaptive mutation over a constant rate (Fig. 4). Consequently, we did not use the adaptive mutation rate in the optimisation experiment, instead taking a constant rate of one (equivalent to one gene mutation per parent).

In the case of the PSO, results were much less sensitive to the σ value (Fig. 4). Lower inertia weight values consistently led to better optimization performance. The best results were achieved with a cognitive coefficient of 0 for the lower σ values (0.05 and 0.1), while for higher σ values (0.25 and 0.5), the cognitive coefficient appeared to be less relevant compared to other tunable parameters. The social coefficient performed better when set above 0.6, with the upper range of high performance around

1.8. Taking into account these results, we decided to use the following values: 0 for the inertia weight, 0.2 for the cognitive coefficient, and 1.4 for the social coefficient.

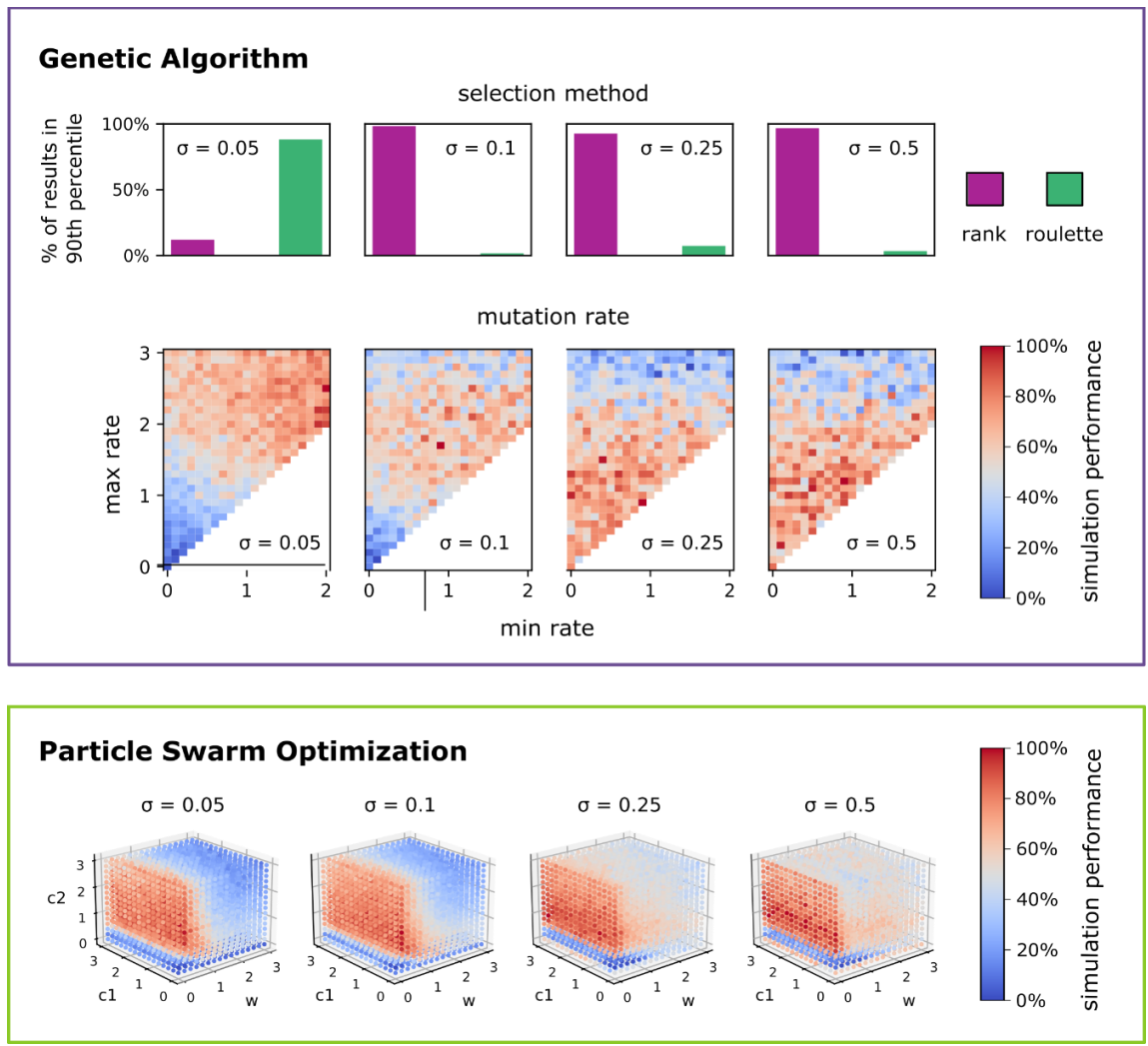

**Fig. 4 Simulation results for the two optimization algorithms: genetic algorithm (GA) and particle swarm optimization (PSO).** Performance of the GA and PSO across different parameter settings after four iterations (one randomized and four processed by the algorithm) using a fitness function modelled as the sum of 1-dimensional normal distributions with maxima at 0.75 of the parameter space range in each dimension. Four sets of simulations were performed, each with a different σ value (0.05, 0.1, 0.25, and 0.5), and 1000 repetitions were executed for each combination of the algorithm parameters. **GA (top)** – performance comparison for rank and roulette selection method, **(bottom)** – the impact of different minimum and maximum mutation rates. The roulette selection method was optimal at σ = 0.05, while for other values, the rank selection method produced the best results. Additionally, as σ increases, a lower mutation rate becomes more advantageous. **PSO** results for the combination of its tunable parameters: inertia (w), cognitive coefficient (c1) and social coefficient (c2), each varied in the range of 0 to 3 with steps of 0.2. Two bottom plot results are normalized to the best optimization outcome recorded within a given σ value. The best optimization results were obtained with low values of w, a broad range of c1, and c2 within the range of 0.8 to 1.8.

**LCE film fabrication**

The LCE films were fabricated in the glass cells, made from two microscopic slides, spin-coated with polyvinyl alcohol (PVA). They were separated with a wire of well-defined diameter, that determined robot thickness. To imprint the director orientation pattern in the LCE films, parts of the PVA layer, unrestricted by a mask, were rubbed in one direction with a velvet cloth, following the method described in [23]. The cell was then filled with the LCE precursor (Supplementary Table 1) by capillary action, using a hot plate set to 80°C. After a passive cool-down period of 15 minutes, the mixture was polymerized with a 379 nm UV LED that triggered a photopolymerization reaction through the activation of the photoinitiator. After opening the cell, the LCE film was cut by hand to the desired size with the help of a scalpel.

**Experimental Setup**

The robot body was a 3 mm wide strip cut from the LCE film. The strip was fully submerged in distilled water and placed between two parallel glass slides that helped to maintain its course during races. A laser beam (Verdi V-5, Coherent, 5.0 W maximum power), focused with a cylindrical lens into an ellipse with full widths of (1.85 ± 0.05) mm and (0.04 ± 0.02) mm at half maximum at the robot position, was used to locally heat the robot through light absorption of the dye (Fig. 1). The light power reaching the robot, measured with a Coherent LabMax-TOP model LM-45 HTD power meter, was determined to be 80% of the power displayed on the laser control panel. The beam was reflected from the mirror mounted onto the galvo scanner, driven with a sawtooth signal and a single scanning cycle consisting of two parts: the slow scan of the laser beam inducing deformation travelling along the robot, and the fast scan in the opposite direction (shorter than 1% of the cycle duration). This mechanism helped move the beam back to the starting position with no noticeable effect on the robot. The robot locomotion was recorded using a CCD camera (DLT-Cam PRO 5 MP, Delta Optical) covered with an absorption filter, to mask the laser light. Movies captured by the camera were later used to determine the robot's speed following the protocol described below.

**Measurements**

We investigated the performance of two optimization algorithms: GA and PSO, utilizing the measurement results to evaluate the fitness functions of both algorithms. The optimization started with randomly selecting the parameters for the first generation of robots. Next, the robots with corresponding properties were fabricated and their swimming speed was measured. This optimization process was conducted iteratively, involving repeated speed measurements and the generation of parameters for the next generation of robots by the algorithm. Following this procedure, for each of the two randomized initial sets of robots, we independently performed two optimizations: one using GA and one using PSO, resulting in four optimization experiments in total. Details of the experiments can be found in the Supplementary information.

# Data availability

The complete dataset and source code used for this research have been deposited on Zenodo and are accessible via DOI: https://doi.org/10.5281/zenodo.15158295.

## Acknowledgments

This work was supported by the National Science Centre (NCN, Poland) with grant No. 2018/29/B/ST7/00192 "Micro-scale actuators based on photo-responsive polymers", awarded to P.W.; by the Polish Ministry of Science and Higher Education (MNiSW) with grant "Diamentowy Grant" project No. DI2016 015046, awarded to M.R.; and by the Polish Ministry of Science and Higher Education (MNiSW) with grant "Perły Nauki" (Project No. PN/01/0158/2022), awarded to Z.D.

We would also like to thank Milena Królikowska and Alexander Krupiński-Ptaszek for their advice and assistance on the programming aspects of this work. Their insightful suggestions and guidance played an important role in development of the software used in this project.

## Author contributions

Conceptualization: MR, ZD, PW; Formal analysis: MR; Funding acquisition: MR, ZD, PW; Investigation: MR, ZD; Methodology: MR, ZD; Project administration: MR; Software: MR; Resources: MR, PW; Supervision: MR, PW; Validation: MR; Visualization: MR, ZD; Writing – original draft: MR, PW; Writing – review & editing: MR, ZD, PW.

## Competing interests

The authors declare that they have no competing interests.

# Supplementary information

## LCE mixture composition

| COMPOUND TYPE | COMPOUND NAME | VENDOR | CONCENTRATION [mol%] | |
|---|---|---|---|---|
| | | | solution 1 | solution 2 |
| MONOMER | ST03866<br>CAS: 130953-14-9 | Synthon Chemicals | 93.0 | 93.8 |
| CROSSLINKER | ST03021<br>CAS: 174063-87-7 | Synthon Chemicals | 5.0 | 5.0 |
| PHOTOINITIATOR | Irgacure 369<br>CAS: 119313-12-1 | Sigma-Aldrich | 1.0 | 1.0 |
| DYE | Disperse Red 1<br>CAS: 2872-52-8 | Sigma-Aldrich | 1.0 | 0.2 |

**Tab. 1 Molar concentration of compounds used to create LCE precursor.** Before each use, the mixture created from the listed compounds was heated to 140°C for 30 minutes to fully melt all of the crystals.

## Robot speed measurement

After preliminary experiments, we observed that at the beginning of the measurement, the laser power was used for the initial heating up of the material, resulting in behaviour that differed significantly from the later stages of the experiment. Therefore, to ensure the consistency of our measurements, we implemented a standardising procedure. Each film was dried and stored out of water for at least 10 minutes before each set of measurements.

Each measurement session began by removing bubbles forming on the surface of the film and a 20-second 'pumping scan', during which the galvo was driven with a symmetrical sawtooth signal, with other parameters such as frequency and laser power set as the target values. Subsequently, we changed the galvo driving signal to an asymmetric mode and started recording a 40-second movie of the robot's motion. If the robot moved fast enough to step out of the camera frame (approximately 5 cm wide) during measurement, we stopped recording before the full 40 seconds had elapsed.

It was observed that the speed of the robot often varied depending on the direction of the laser scan, reflecting the robot's geometry. To account for this variability, each robot was measured 10 times: five measurements were conducted with slow scans in one direction, and the remaining five with slow scans in the opposite direction. The absolute value of the average slope factor was then calculated for each robot's measurements, separately for each direction of the slow scan (yielding two values per robot). We assumed that the robot's speed was the maximum of these two values.

## Results of the speed optimization - second experiment

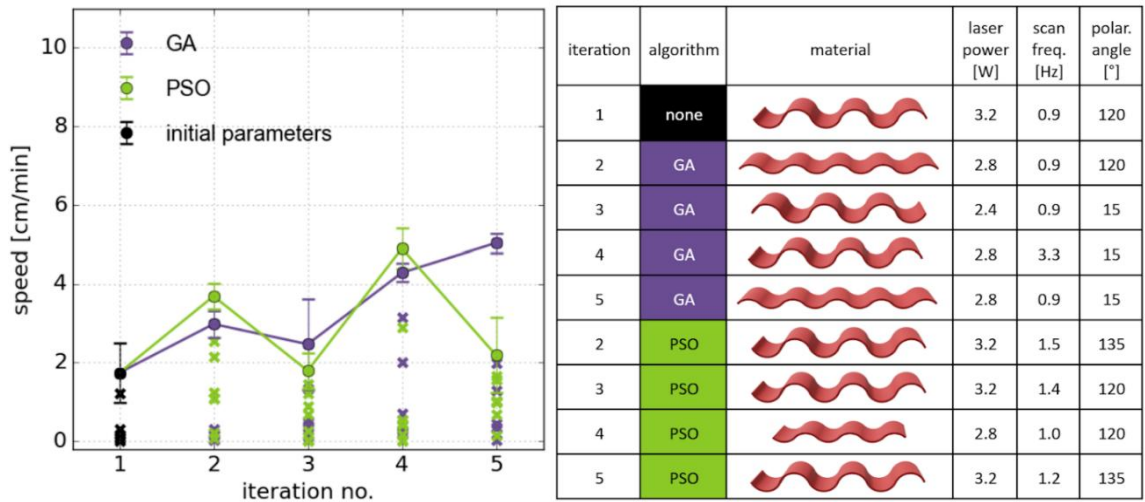

**Fig. 1 Results of the second repetition of the swimming microrobot optimisation process. (Left)** measured speed of the fastest robot in each iteration after evolution performed by the genetic algorithm (GA) and by the particle swarm optimisation (PSO). Bars represent the standard deviation of the results for five measurements. (**Right**) A table summarising the parameters of the fastest swimming microrobots achieved in each iteration by both optimisation methods. Material parameters are illustrated on 3D models, while the laser beam parameters (including laser power, scanning frequency, and the angle between the laser polarisation and the robot's body) are detailed in the three right-hand columns.